\documentclass[11pt]{article}

\usepackage{acl}

\usepackage{times}
\usepackage{latexsym}
\usepackage{float}
\usepackage{colortbl}
\usepackage{tabularx}
\usepackage{longtable}
\usepackage{xcolor} 
\usepackage{booktabs}

\usepackage{microtype}
\usepackage{booktabs}
\usepackage{multicol,multirow}

\usepackage{inconsolata}

\usepackage{graphicx}
\usepackage{float}
\usepackage{amsfonts,amsmath,amssymb,amsthm}
\usepackage[ruled,linesnumbered]{algorithm2e}
\usepackage{soul}

\makeatletter
\renewcommand{\@fnsymbol}[1]{$\dagger$}
\makeatother

\title{Improving Multilingual Instruction Finetuning \\
via Linguistically Natural and Diverse Datasets}

\author{Sathish Reddy Indurthi\thanks{Correspondence to \texttt{sathishreddy.indurthi@zoom.us}}, Wenxuan Zhou, Shamil Chollampatt, Ravi Agrawal \\
\textbf{Kaiqiang Song, Lingxiao Zhao, Chenguang Zhu}\\
Zoom Video Communications}

\begin{document}
\maketitle
\begin{abstract}

Advancements in Large Language Models (LLMs) have significantly enhanced instruction-following capabilities. However, most Instruction Fine-Tuning (IFT) datasets are predominantly in English, limiting model performance in other languages. Traditional methods for creating multilingual IFT datasets—such as translating existing English IFT datasets or converting existing NLP datasets into IFT datasets by templating—struggle to capture linguistic nuances and ensure prompt (instruction) diversity. To address this issue, we propose a novel method for collecting multilingual IFT datasets that preserves linguistic naturalness and ensures prompt diversity. This approach leverages English-focused LLMs, monolingual corpora, and a scoring function to create high-quality, diversified IFT datasets in multiple languages. Experiments demonstrate that LLMs finetuned using these IFT datasets show notable improvements in both generative and discriminative tasks, indicating enhanced language comprehension by LLMs in non-English contexts. Specifically, on the multilingual summarization task, LLMs using our IFT dataset achieved 17.57\% and 15.23\% improvements over LLMs fine-tuned with translation-based and template-based datasets, respectively.

\end{abstract}

\section{Introduction}
Recent advancements in natural language processing (NLP) have showcased remarkable progress, particularly in its instruction-following capabilities. Notably, Large Language Models (LLMs) like GPT-4, Gemini-1.5, Claude-3, Llama-3, and Mistral \cite{gpt-4, geminiteam2024gemini, llama3modelcard, jiang2023mistral} have demonstrated significant prowess in this area \cite{brown2020language, workshop2023bloom, chowdhery2023palm}. After the pretraining stage, LLMs are fine-tuned on Instruction Fine-Tuning (IFT) datasets followed by an optional Alignment Tuning (AT) based on the availability of the training datasets. IFT datasets consist of instruction prompt-response pairs and have proven instrumental in enhancing the efficacy and overall instruction-following abilities of LLMs \cite{anil2023palm, sanh2022multitask, wei2023polylm, iyer2023optiml, chung2022scaling, wang2022supernaturalinstructions, zhang2024instruction}. However, a notable disparity persists between the abundance of instruction prompts available in English compared to other languages. While over 7k\footnote{https://www.ethnologue.com/} languages are spoken worldwide, a staggering 73\% of prevalent IFT datasets primarily cater to English alone \cite{longpre2023data}.

\begin{figure}[t]
    \centering
   \includegraphics[width=1.0\linewidth]{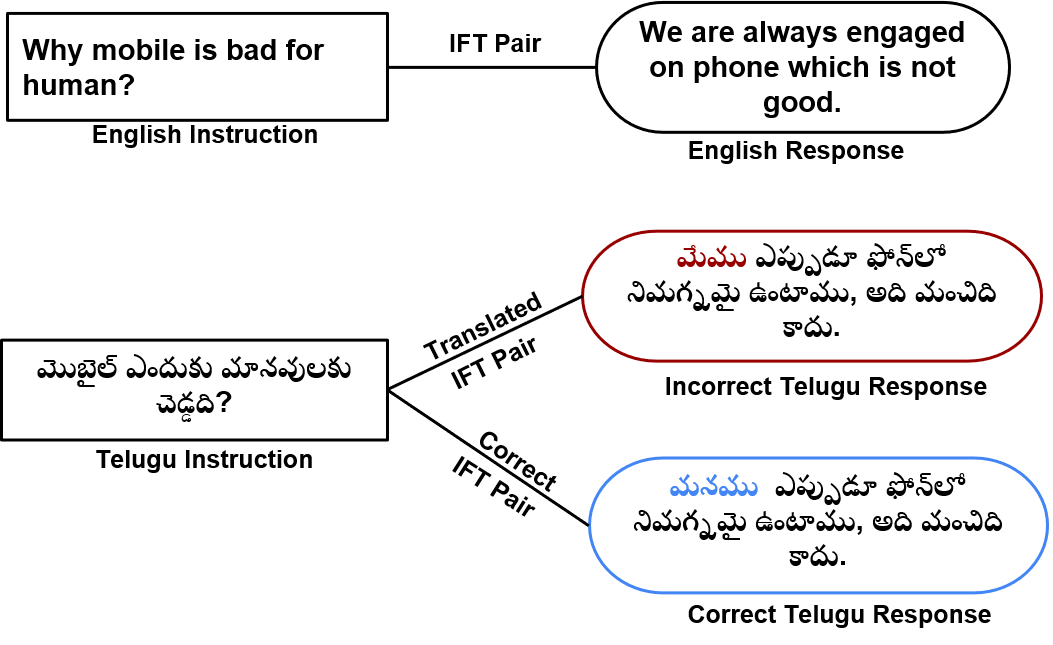}
    \caption{The incorrectly translated Telugu instruction-response pair is from the Aya collection \cite{2024aya}, which was translated from an English instruction-response pair in the Dolly v2 dataset \cite{DatabricksBlog2023DollyV2}. The correct Telugu instruction-response pair was provided by a native Telugu speaker.}
    \label{fig:tr_er_ex}
\end{figure}

While LLMs often demonstrate proficiency in understanding and generating text across multiple languages, the language imbalance in training datasets has led to suboptimal performance in non-English contexts \cite{ahuja2023mega, lai2023chatgpt, zhang2023dont}. To enhance LLMs' ability to follow non-English instructions, various studies have explored fine-tuning LLMs on multilingual Instruction Fine-Tuning (IFT) datasets \cite{muennighoff2023crosslingual, wei2023polylm, lai2023okapi, zhang2024plug, shaham2024multilingual, chen2024monolingual,2024aya}. However, creating such multilingual IFT datasets is challenging. Previous efforts have primarily focused on two approaches: translating existing English IFT datasets or templating existing Natural Language Processing (NLP) datasets in non-English languages through native speakers to form IFT-style datasets. Each approach has its drawbacks, highlighting the need for more effective methods.

Translating English IFT datasets poses significant challenges, primarily because it fails to capture the nuances and intricacies unique to each language \cite{liu2024translation, zhang2023m3exam}. Additionally, the translation process often introduces errors, leading to suboptimal performance when fine-tuning LLMs on these translated datasets, as the models absorb these errors during training \cite{xu2023wizardlm, zhou2023lima, pmlr-v202-kong23a}. For example, in Figure \ref{fig:tr_er_ex}, the first translated response (red) is incorrect, even though it differs from the correct response (blue) by just one word. The red and blue words are used in different contexts in Telugu and do not have direct translations in English. Despite being generated by a state-of-the-art translation model, the first translation (red) fails to capture the correct meaning. Thus, relying entirely on translated data poses significant challenges in accurately reflecting the nuances of non-English languages.

\begin{figure}
    \centering
    \includegraphics[width=1.0\linewidth]{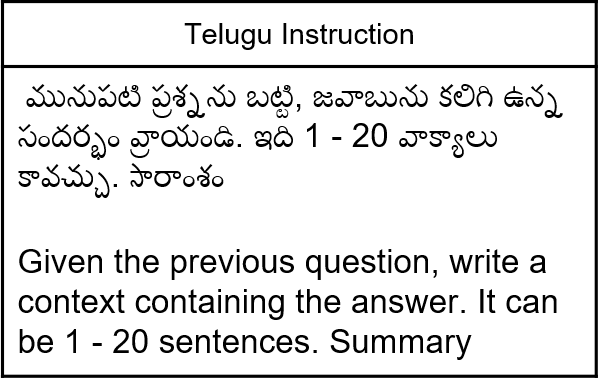}
    \caption{Lack of diversity in templated datasets: The template created by human annotators has been repeated several thousand times in the templated adversarial QA dataset from the Aya collection \cite{2024aya}}
    \label{fig:tp_er_ex}
\end{figure}

Comparatively, the templating approach avoids the introduction of translation errors. However, achieving high diversity through templated approaches is challenging and often tedious due to the manual effort required \cite{muennighoff2023crosslingual, sanh2022multitask}. For instance, as shown in Figure \ref{fig:tp_er_ex}, one of the templated datasets contains the same instruction repeated several thousand times, resulting in a lack of diversity in the IFT dataset.

To address the issues of translation and templated approaches, we introduce an efficient method to collect high-quality multilingual IFT datasets. The proposed method preserves the nuances of languages, avoids errors, and creates a diverse set of IFT examples for multiple languages. This is achieved by leveraging an English-focused LLM and the availability of monolingual corpora in each non-English language. We also employ a scoring function to control the quality of generated IFT examples. By relying on English-focused LLMs, we can tap into their extensive capabilities and transfer these abilities across diverse linguistic contexts. Utilizing monolingual corpora allows us to capture the unique linguistic and cultural nuances of each language, enhancing performance and accuracy in multilingual applications. Additionally, the robust scoring function ensures that the knowledge and capabilities derived from English-centric LLMs are appropriately adapted and optimized for non-English languages.

Extensive experiments on both generative and discriminative tasks demonstrate the effectiveness of the multilingual IFT datasets resulting from our proposed method. Compared to models fine-tuned on IFT datasets created using translation and templated approaches, the model fine-tuned on IFT datasets from our method achieves an average improvement of 11.1\% in generative tasks and 6.9\% in discriminative tasks. Furthermore, these improvements are obtained with an IFT dataset less than half the size of those created using templated and translation methods, highlighting the superior quality and diversity of the IFT dataset generated by our approach.

\begin{figure*}[t]
\centering
\includegraphics[width=1.0\linewidth]{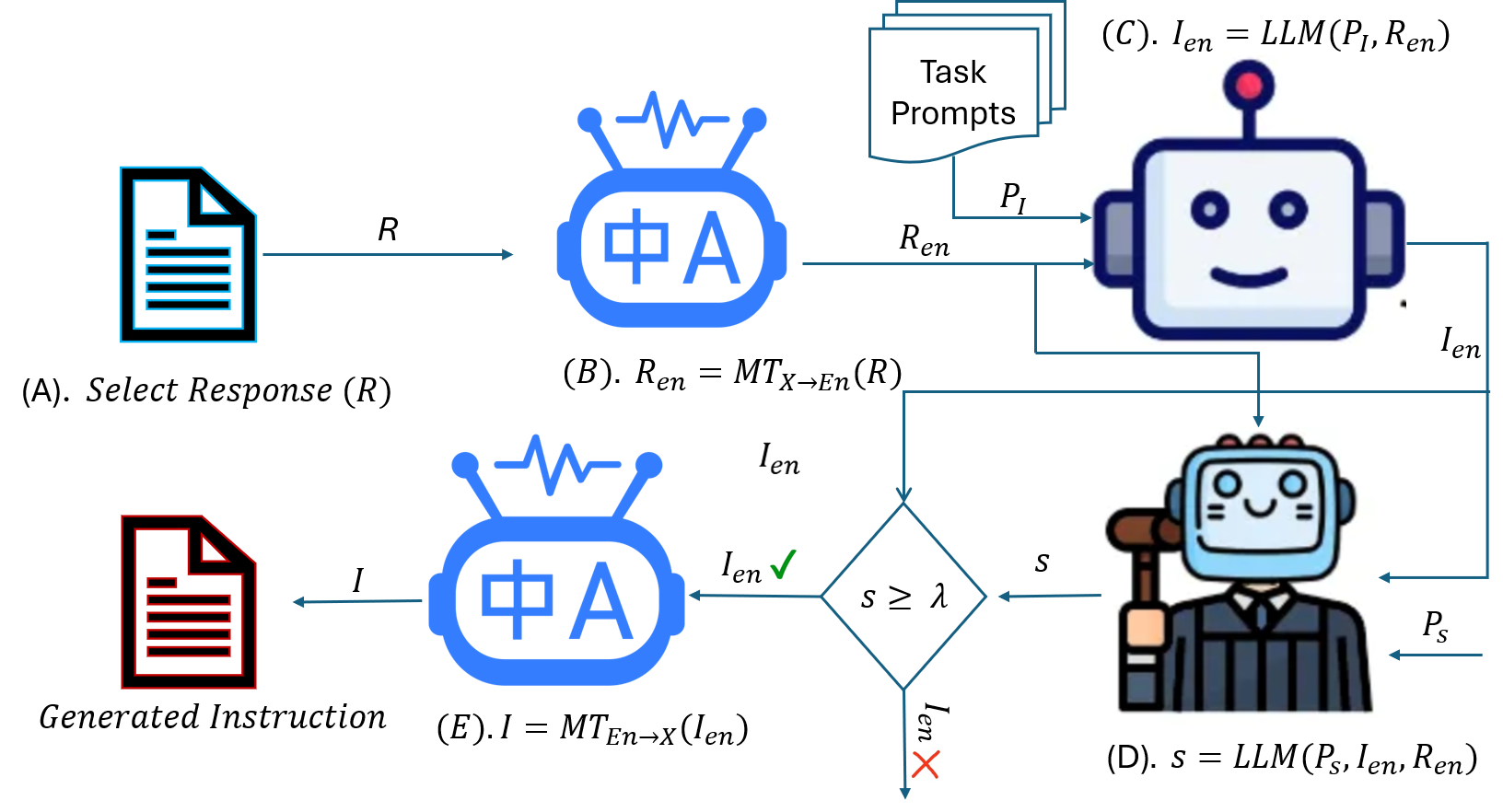}
\caption{Overview of the proposed method: (A) Select Response, (B) Translating Response to English, (C) Generating English instructions using the English Response and task-specific prompt, (D) Scoring the generated English instruction against the translated response, and (E) Translating the English instruction back to the language of the original response.}
\label{fig:method}
\end{figure*}

\section{Method}
\label{sec:method}
A fundamental component in the development of Multilingual Large Language Models (MLLMs) lies in the acquisition of training datasets, crucially needed throughout distinct phases: Pretraining (PT), Instruction fine-tuning (IFT), and Alignment Tuning (AT).

While obtaining the necessary monolingual datasets for pretraining is relatively straightforward, acquiring datasets for instruction fine-tuning and alignment tuning presents significant challenges due to the costs and human effort involved. To address these challenges while maintaining linguistic characteristics and diversity, we propose a framework for creating IFT datasets for multiple languages. The framework consists of five stages, illustrated in Figure \ref{fig:method} and described below:

\paragraph{(A). Select Responses:} 
We utilize a monolingual corpus as the primary source of response, supplemented by answers from existing NLP datasets for each non-English language $(x)$. We extract several thousand text fragments from these non-English corpora, deduplicate, and apply various heuristics to filter out potentially low-quality fragments. These heuristics include criteria such as the prevalence of capitalized letters and specialized symbols. These text fragments are natural and most likely error-free output since they are from the monolingual corpus or human-curated answers from existing NLP datasets. Each fragment, denoted as $\mathcal{R}_{x}$, which varies in length to resemble responses in real-world scenarios, is then used to generate pseudo instructions through the following steps. By doing this, we ensure the output quality of the multilingual IFT data.

\paragraph{(B). Translating Responses:}
Given the availability of competent English LLMs in both open-source and closed environments, we have chosen to generate pseudo-instructions in English. This strategic decision allows us to leverage the strength of these models, ensuring the generation of high-quality and diverse instructions that cater to a wide range of NLP tasks, we translate the selected response ($\mathcal{R}_{x}$) into English.

\begin{equation*}
    \mathcal{R}_{en} = \mathcal{MT}_{x \rightarrow en}(\mathcal{R}_{x}) 
\end{equation*}

\paragraph{(C). Generating Instruction:}
We generate English instructions by utilizing English-focused LLM, a translated response ($\mathcal{R}_{en}$), and a randomly selected prompt ($\mathcal{P}_{I}$) from a pool of predefined task prompts. Our approach involves designing a range of prompts specifically tailored to support various NLP tasks, including question-answering, summarization, and sentiment analysis. Additionally, the prompt allows for open-ended instruction generation, providing LLMs with the opportunity to produce the most plausible instructions for a given response. Focusing on generating instructions in English enables us to tap into the extensive resources and capabilities available for this language, thereby enhancing the adaptability and effectiveness of our approach across diverse linguistic contexts. This emphasis on English instruction generation also ensures seamless integration with existing English-centric NLP systems, further augmenting the versatility and applicability of our methodology in real-world scenarios. Formally, the English instruction ($\mathcal{I}_{en}$) is generated by:
\begin{equation}
    \mathcal{I}_{en} = \mathcal{LLM}(\mathcal{P}_{I}, \mathcal{R}_{en}) 
    \label{eq:instruct_gen}
\end{equation}
    
\paragraph{(D). Scoring}
The instructions generated through (\ref{eq:instruct_gen}) do not always yield high-quality examples due to misalignment in the prompt-response pair or LLM's failure to generate appropriate instruction. Thus we rely on a scoring function to filter and identify high-quality examples while maintaining diversity in the generated dataset.

We use LLM as a judge, employing the prompt $\mathcal{P}_{s}$ to assess the quality of $(\mathcal{I}_{en}, \mathcal{R}_{en})$ pair. This results in a score, denoted as $s$:
\begin{equation}
s = \mathcal{LLM}(\mathcal{P}_{s}, \mathcal{I}_{en}, \mathcal{R}_{en})
\label{eq:score}
\end{equation}

Pairs with a score greater than or equal to a predefined threshold ($\lambda$) are used for fine-tuning, while those below this threshold are removed from fine-tuning phase.

\paragraph{(E). Translating Instruction:}

Following the scoring phase, we proceed to translate $\mathcal{I}_{en}$ into the same language as $\mathcal{R}_{x}$:
\begin{equation*}
    \mathcal{I}_{x} = \mathcal{MT}_{en \rightarrow x}(\mathcal{I}_{en}) 
\end{equation*}

Subsequently, we form a training pair $(\mathcal{I}_{x}, \mathcal{R}_{x})$. Here, $\mathcal{I}_{x}$ serves as a pseudo instruction, while $\mathcal{R}_{x}$ represents natural text in the same non-English language. During the LLM fine-tuning stage, despite potential unnaturalness and errors in $\mathcal{I}_{x}$ arising from the instruction generation and translation process, the model is trained to generate $\mathcal{R}_{x}$, which is typically a natural and error-free output sourced from the monolingual corpus or existing human-curated NLP datasets. Leveraging such pairs enhances the model's ability to handle instruction errors and improves its overall language comprehension.

The sample task prompts ($\mathcal{P}_{I}$) and scoring prompt ($\mathcal{P}{s}$) used in Equation \ref{eq:instruct_gen} and Equation \ref{eq:score} are provided in Table \ref{tab:sample_task_prompts} and Table \ref{tab:score_prompt} in the Appendix.

\section{Experimental Settings}
\subsection{Dataset creation}

\begin{table}[t]
    \centering
    \begin{tabular}{|c|c|c|c|} \hline
         Language &  TM & TR  & GR \\ \hline
         Telugu & 1,312,185 & 2,596,857 & 523,739 \\
         Hindi & 1,171,530 & 2,540,447 & 570,467 \\
         Japanese & 2,392,691 & 3,029,014 & 531,163 \\
         Spanish & 1,220,649 & 2,560,149 & 557,563\\ \hline
    \end{tabular}
    \caption{Total number of instruction-response pairs used for fine-tuning the LLMs by Templated (TM), Translation (TR), and Generated (GR) approaches.}. 
    \label{tab:data_stats2}
\end{table}

\begin{table*}[t]
    \centering
    \begin{tabular}{|c|c|c|c|c|c|c|} \hline
     \multirow{2}{*}{\textbf{Language}}  & \multicolumn{2}{c|}{\textbf{Templated}}  &  \multicolumn{2}{c|}{\textbf{Translated}} & \multicolumn{2}{c|}{\textbf{Generated}} \\
                
            &    Instruction & Response & Instruction & Response & Instruction & Response \\ \hline
                                            
         Telugu (tel) & $344 (\pm312)$ & $221 (\pm297)$ & $223 (\pm295)$ & $204 (\pm179) $ & $381 (\pm917)$ & $308 (\pm482)$  \\
          Hindi (hin) & $401 (\pm450)$ & $290 (\pm315)$ & $228 (\pm397)$ & $203 (\pm181)$ & $475 (\pm897)$ & $358 (\pm582)$ \\
         Japanese (jpn) & $67 (\pm79)$ & $95 (\pm115)$ & $94 (\pm172)$ & $86 (\pm78)$  & $162 (\pm473)$ & $98 (\pm116)$\\
         Spanish (spa) & $306 (\pm280)$ & $138 (\pm215)$  &  $238 (\pm435)$ & $215 (\pm196)$ & $425 (\pm 723)$ & $289 (\pm475)$\\ \hline
    \end{tabular}
    \caption{Average lengths (\#characters) of instruction-response pairs in templated, translated, and generated approaches.}
    \label{tab:data_stats1}
\end{table*}

We utilize the CC-100 monolingual dataset \cite{conneau2020unsupervised}. We also utilize answers from the templated examples in the aya dataset \cite{2024aya}.  In both cases, the texts are written in the native language not derived from other languages \cite{wenzek2020ccnet}. We selected the text based on the criteria described in Section \ref{sec:method}. We choose four languages: Telugu (tel), Hindi (hin), Japanese (jpn), and Spanish (spa) to create IFT datasets through our approach. According to Aya and Okapi \cite{2024aya, lai2023okapi}, Telugu and Nepali are low-resource, Indonesian and Hindi are mid-resource, and Japanese and Spanish are high-resource languages. We collected approximately one million text fragments for each language. 

In creating multilingual datasets using the proposed approach, we utilize open source \textit{meta-llama/Meta-Llama-3-70B-Instruct} \cite{llama3modelcard} as our LLM to generate instructions and also to score instruction-response pairs. However, this LLM can be replaced with more powerful open-source or closed-source LLMs to improve the quality of generated instructions further. 

We utilize NLLB-200 \cite{nllb2022}\footnote{https://huggingface.co/facebook/nllb-200-3.3B}, which has support for 200 languages with state-of-the-art translation quality. The same model is used for translating the response ($\mathcal{R}$) to English as well as for translating ($\mathcal{I}_{en}$) into the language of ($\mathcal{R}$).
After the translation, we use the COMET score \cite{rei2020comet} to remove low-quality translated responses $(\mathcal{R}_{en})$ and generated instructions $(\mathcal{I}_{x})$. Specifically, we use \textit{Unbabel/wmt23-cometkiwi-da-xl} model \cite{rei2023scaling}, which is a reference-free model with 3.5 billion parameters. We retain examples with COMET scores greater than or equal to $0.7$.

\subsection{Training details}
We use 
\textit{Meta-Llama-3-8B} \cite{llama3modelcard}\footnote{https://huggingface.co/meta-llama/Meta-Llama-3-8B} as the base model to fine-tune on our multilingual IFT dataset. We also fine-tune non-English focused models such as  \textit{Rakuten-ai-7B-Instrcut} \cite{rakutengroup2024rakutenai7b}, \textit{Aya-23} \cite{aryabumi2024aya}. During training, we only optimize the loss on the output tokens, not the input tokens, thus deviating from the standard language modeling loss. We apply the same hyperparameters as existing instruction fine-tuning (IFT) methods \cite{zhou2023lima, touvron2023llama}: a learning rate of $1e^{-5}$ that linearly decays to $9e^{-6}$ by the end of the training, weight decay of 0.1, batch size of 128 examples, and dropout of 0.1. For generation, we use nucleus sampling \cite{holtzman2020curious} with a temperature of $T = 0.7$ and $p = 0.9$. We use 8 NVIDIA H100 GPUs for fine-tuning the model.

\section{Results}

In Table \ref{tab:data_stats2}, we present the statistics of datasets created using various approaches. The statistics of datasets created using the template-based and translation-based approaches are from \textit{aya\_collection} \cite{2024aya}. Please see the Appendix for more details. Using our approach, we generated approximately 500K instruction-response pairs from the initial pool of 1M text fragments for each language. 

We evaluate the performance of models fine-tuned on datasets collected using our approach against models fine-tuned on datasets obtained through translation and template-based methods. Specifically, we compare the \textit{Aya-TM} and Llama-3-8B-TM models, which are trained on template-based datasets as described in \citet{2024aya}. Additionally, we assess the \textit{Aya-TR} and Llama-3-8B-TR models, which are trained on translation-based datasets detailed in \citet{2024aya}. Both types of datasets include the Aya-human annotated dataset\footnote{https://huggingface.co/datasets/CohereForAI/aya\_dataset}. Furthermore, we compare these with the \textit{Bactrian-X} model \cite{li2023bactrian}, fine-tuned on a dataset comprising translated English instructions and their corresponding multilingual responses generated using ChatGPT. Our final model, Llama-3-8B-GR, is trained using the created instruction-response dataset along with the Aya human-annotated dataset. In all the approaches, the percentage of training examples collected through the human annotation process corresponds to less than $0.1\%$. 

\begin{table}[t]
\centering
\begin{tabular}{lrrrr}
\\ \hline
    \multicolumn{5}{c}{\textbf{RougeLsum}} \\ \hline
  & \multicolumn{1}{l}{\textbf{tel}} & \multicolumn{1}{l}{\textbf{hin}} & \multicolumn{1}{l}{\textbf{jpn}} & \multicolumn{1}{l}{\textbf{spa}}\\ \hline
\multicolumn{5}{c}{Templated Approaches} \\ \hline
Aya-TM   & 18.0   &  33.8 & 7.9  &  24.2  \\
Llama-3-8B-TM  & 19.6 & 36.4 & 17.8 & 26.8  \\ \hline
\multicolumn{5}{c}{Translated Approaches} \\ \hline
Bactrian-X & 12.1 & 23.5 & 5.2 & 15.7 \\
Aya-TR   & 16.9   &  32.8  &  6.7  & 22.1 \\ 
Llama-3-8B-TR &  18.4 & 35.9 & 18.4 &  25.9 \\ \hline
\multicolumn{5}{c}{Ours} \\ \hline
Llama-3-8B-GR  & \textbf{24.3} & \textbf{39.5}  & \textbf{22.6}  & \textbf{29.5} \\ \hline
\end{tabular}
\caption{Performance of models on XLSUM.}
\label{tab:res_xlsum}
\end{table}

\begin{table}[t]
\centering
\begin{tabular}{lrrrr}
\\ \hline
 \multicolumn{5}{c}{\textbf{spBleu}}
\\ \hline
  & \multicolumn{1}{l}{\textbf{tel}} & \multicolumn{1}{l}{\textbf{hin}} & \multicolumn{1}{l}{\textbf{jpn}} & \multicolumn{1}{l}{\textbf{spa}}\\ \hline

\multicolumn{5}{c}{Templated datasets} \\ \hline
Aya-TM            &  21.9 & 22.7 & 18.2 & 27.1    \\
LLama-3-8B-TM       &  24.6 &  25.3 & 21.6 &  30.7 \\ \hline

\multicolumn{5}{c}{Translated datasets} \\ \hline
Bactrian-X & 17.3 & 19.2 & 11.78 & 22.4 \\
Aya-TR              &  21.0 & 22.8 & 14.7 &  28.4  \\
Llama-3-8B-TR     &  23.5 &  24.9 & 20.2 & 31.2 \\ \hline

\multicolumn{5}{c}{Ours} \\ \hline
Llama-3-8B-GR      & \textbf{27.2} & \textbf{28.4} & \textbf{24.8} & \textbf{33.9}  \\ \hline
\multicolumn{5}{c}{\textbf{chrF++}}
\\ \hline
\multicolumn{5}{c}{Templated datasets} \\ \hline
Aya-TM   & 44.7 & 44.1  &  29.7 & 50.3   \\
Llama-3-8B-TM  &  47.1  &  46.9  &  34.7  &    58.4  \\ \hline

\multicolumn{5}{c}{Translated datasets} \\ \hline
Bactrian-X & 35.8 & 36.9 & 22.1 & 42.8 \\
Aya-TR   & 45.5 & 44.9  &  29.9  & 51.9  \\ 
Llama-3-8B-TR  &  47.7  & 46.4  &  35.0  &    57.7  \\ \hline

\multicolumn{5}{c}{Ours} \\ \hline
Llama-3-8B-GR  &  \textbf{49.8}  & \textbf{50.2}  &  \textbf{38.3}  & \textbf{63.2}      \\ \hline
\end{tabular}
\caption{Performance of models on FLORES-200 devtest set (en$\rightarrow$xxx).}
\label{tab:res_flores200}
\end{table}

\subsection{Generative Tasks}
We evaluated the models on two generative tasks: summarization using \textit{XLSUM} \cite{hasan2021xlsum} and machine translation using \textit{FLORES-200} \cite{nllb2022}. These tasks were selected because they include responses written in native languages, not derived from other languages. We present the performance of our model, Llama-3-8B-GR-H, and its variants, comparing them to baseline models across the four languages used for creating multilingual IFT datasets. For the summarization task, we employed the RougeLsum metric \cite{lin-2004-rouge}, and for the translation task, we utilized spBLEU \cite{spBleu} and chrF++ \cite{popovic-2017-chrf}\footnote{https://github.com/mjpost/sacrebleu}.

Tables \ref{tab:res_xlsum} and \ref{tab:res_flores200} present the results for the summarization and machine translation tasks using the XLSUM and FLORES-200 datasets, respectively. From the results presented in both tables, models trained with translated datasets do not exhibit any improvement over those trained with template datasets. In contrast, the Llama-3-8B-GR model, fine-tuned on datasets created using our method, demonstrates significant performance enhancements across both tasks compared to all other dataset types. Our dataset, free from translation errors and rich in diversity, enables the model to better capture the authentic form of language, leading to superior performance.

\subsection{Discriminative Tasks}

\begin{table}[t]
\centering
\begin{tabular}{lrrr}
\\ \hline
  & \multicolumn{1}{l}{\textbf{tel}} & \multicolumn{1}{l}{\textbf{hin}}  & \multicolumn{1}{l}{\textbf{spa}}\\ \hline
\multicolumn{4}{c}{Translated datasets} \\ \hline
Bactrian-X & 24.5 & 26.2 & 27.2 \\
Okapi            &  26.9   &  27.9  &  30.3  \\
Aya-TR        &  32.1   &  38.7  &  39.7  \\ 
Llama-3-8B-TR  &  34.1 &  41.4  &   42.9     \\ \hline
\multicolumn{4}{c}{Ours} \\ \hline
Llama-3-8B-GR  &  \textbf{36.3}   &  \textbf{44.7}  &     \textbf{45.6}   \\ \hline
\end{tabular}
\caption{Performance of models on multilingual MMLU task.}
\label{tab:res_mmlu}
\end{table}

\begin{figure}[t]
    \centering
    \vspace{-1cm}
    \includegraphics[width=1.0\linewidth]{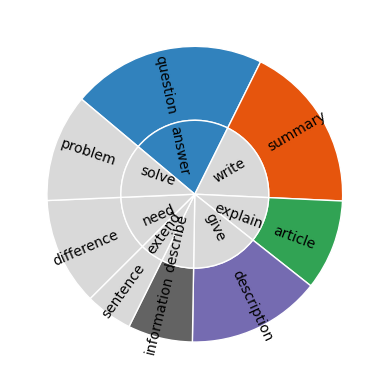}
    \caption{Instruction diversity in the generated IFT dataset. The inner circle displays common root verbs, while the outer circle shows the corresponding noun objects, based on approximately 15 percent of instructions generated across 4 languages. The figure only represents 13.1\% of verb-noun pairs since not all instructions have the parsed verb-noun structure. }
    \label{fig:data_diversity}
\end{figure}

We also evaluate the models on a discriminative task to assess whether introducing high-quality, diversified, and native-written responses enhances the model's language comprehension and overall performance. Specifically, we use the multilingual MMLU task \cite{lai2023okapi}, a machine-translated version of the English MMLU task \cite{hendrycks2020measuring}, to compare the performance of models trained extensively on translated datasets versus those trained on native datasets created using our approach. This task was unseen during the models' fine-tuning stage, so we employ a few-shot evaluation to compare performance. The \textit{Llama-3-8B} and \textit{Aya} models use a 5-shot evaluation, while the \textit{Bactrian-X} and \textit{Okapi} models use a 25-shot evaluation. The task comprises 13,000 questions covering 57 topics, ranging from STEM and humanities to social sciences.

Table \ref{tab:res_mmlu} shows the multilingual results in three languages. The model trained with our dataset (\textit{Llama-3-8B-GR}), outperforms the models trained with datasets collected using other approaches. Our model outperforms Okapi, Aya, and our baseline by 48.74\%, 13.8\%, and 6.9\%, respectively. These results indicate that the diversity and quality of the datasets lead to better performance.

Despite our dataset being 2.7 and 4.9 times smaller than the templated and translated datasets, respectively, the model fine-tuned on our dataset achieved significant improvements in both generative and discriminative tasks. This underscores the importance of high-quality, diversified datasets in developing efficient multilingual LLMs.

\subsection{Analysis}

\subsubsection{Instruction diversity}
To understand the diversity of the generated instructions, we plot the verb-noun structure of instructions in Figure \ref{fig:data_diversity}. The figure visualizes the distribution of the most frequent root verbs and their corresponding most common direct noun objects from 15\% of the generated instructions across four languages. These noun-verb pairs represent $8.1\%$ of the entire set, which exhibits diverse intents and patterns in our generated instructions. We also provide a few generated samples in the Appendix.

We also report the average length of instructions and responses from all data creation approaches. As shown in Table \ref{tab:data_stats1}, the average number of characters in the instructions generated using our approach varies significantly compared to the other two approaches. This variation arises from using different types of task prompts when generating an instruction for a given response.

\subsubsection{Effect of Scoring Function:}

\begin{figure}[t]
    \centering
    \includegraphics[width=1.0\linewidth]{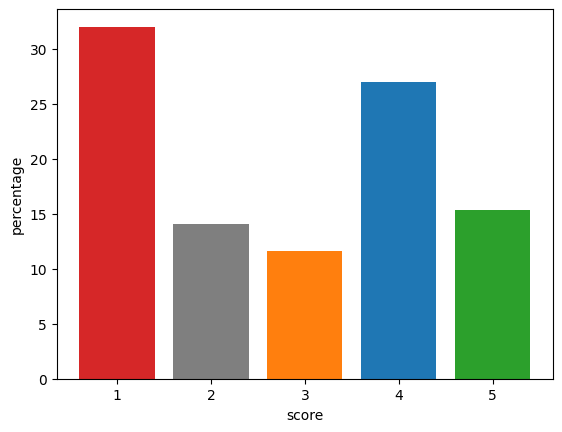}
    \caption{Scores assigned by LLM judge on Instruction-Response pairs. The scores are averaged across all languages.}
    \label{fig:score-stats}
\end{figure}

\begin{figure}[t]
    \centering
    \includegraphics[width=1.0\linewidth]{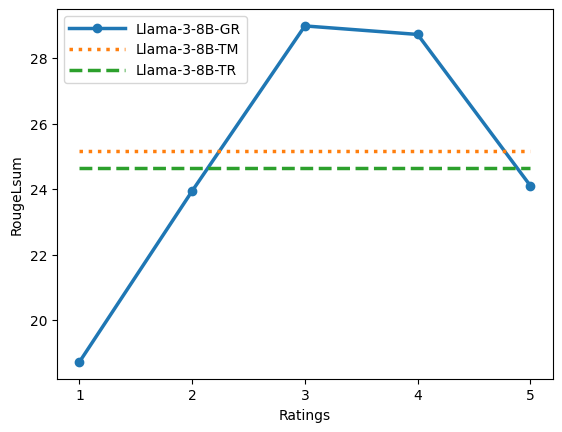}
    \caption{Importance of scoring function in creating high-quality IFT dataset. The x-axis represents the scoring threshold used to filter the IFT dataset. The Y-axis represents the average RougeLsum score of  Telugu, Hindi, Japanese, and Spanish languages from the XLSUM summarization task.}
    \label{fig:score-affect}
\end{figure}

The frequency of average scores obtained using the LLM judge is shown in Figure \ref{fig:score-stats}. To evaluate the impact of the scoring function on the creation of high-quality multilingual IFT datasets, we fine-tuned the \textit{Llama-3-8B-GR} model on datasets filtered using different scoring thresholds, $\lambda = \{1, 2, 3, 4, 5\}$. For each specific threshold $\lambda_i$, all examples below that score were excluded from the training set. We then compared the performance of the \textit{Llama-3-8B-GR} model trained on these filtered datasets against models (\textit{Llama-3-8B-TM} and \textit{Llama-3-8B-TR}) trained on template-based and translation-based datasets. As illustrated in Figure \ref{fig:score-affect}, the performance of \textit{Llama-3-8B-GR} improves as the scoring threshold increases up to $\lambda = 3$, achieving superior performance compared to the \textit{Llama-3-8B-TM} and \textit{Llama-3-8B-TR} models. Beyond $\lambda = 3$, performance declines due to the reduced size of the training dataset. These results underscore the critical role of the scoring function in creating high-quality multilingual IFT datasets.

\subsubsection{Effect on non-English focused models.}

\begin{table}[t]
\centering
\begin{tabular}{lrr}
\hline
\multirow{2}{*}{Model} & \multicolumn{1}{l}{XLSUM} & \multicolumn{1}{l}{MMLU} \\ 
                       & \multicolumn{1}{l}{(Rouge-2)} & \multicolumn{1}{l}{(Acc.)} \\ \hline

\multicolumn{1}{l}{RakutenAI-7B} & \multirow{2}{*}{14.1} & \multirow{2}{*}{61.3} \\
\multicolumn{1}{l}{\small\cite{rakutengroup2024rakutenai7b}} & & \\ \hline
\multicolumn{1}{l}{RakutenAI-7B-GR} & \multirow{2}{*}{\textbf{18.5}} & \multirow{2}{*}{\textbf{63.2}} \\
\multicolumn{1}{l}{(w/ our IFT dataset)} & & \\ \hline

\end{tabular}
\caption{Performance of Japanese-focused LLMs on XLSUM and MMLU Japanese tasks.}
\label{tab:res_japanese}
\end{table}

\begin{table}[t]
\centering
\begin{tabular}{lrr}
\hline
\multirow{2}{*}{Model} & \multicolumn{1}{l}{XLSUM} & \multicolumn{1}{l}{MMLU} \\ 
                       & \multicolumn{1}{l}{(RougeL)} & \multicolumn{1}{l}{(Acc.)} \\ \hline

\multicolumn{1}{l}{Aya-23-8B} & \multirow{2}{*}{29.7} & \multirow{2}{*}{45.3} \\
\multicolumn{1}{l}{\small\cite{aryabumi2024aya}} & & \\ \hline
\multicolumn{1}{l}{Aya-23-8B-GR} & \multirow{2}{*}{\textbf{31.4}} & \multirow{2}{*}{\textbf{46.8}} \\
\multicolumn{1}{l}{(w/ our IFT dataset)} & & \\ \hline

\end{tabular}
\caption{Performance of Aya-23-8B LLM on XLSUM and MMLU Hindi and Spanish tasks. The \textit{Aya-23-8B-GR} model is obtained by further finetuning of \textit{Aya-23-8B} model on our Hindi and Spanish IFT datasets.}
\label{tab:res_aya23}
\end{table}

To evaluate the diversity and quality of our IFT datasets, we conducted further fine-tuning on two robust non-English-focused LLMs using our IFT datasets. First, we assessed the impact on the Japanese-focused model \cite{rakutengroup2024rakutenai7b}. This model was initially pre-trained on Japanese texts and fine-tuned on Japanese instruction-response pairs. Second, we evaluated the performance of a state-of-the-art multilingual LLM named Aya-23 \cite{aryabumi2024aya}. This model is based on Cohere's Command model\footnote{https://cohere.com/command} and was instruction-tuned on 23 languages using the template-based dataset from \citet{2024aya}.

As shown in Table \ref{tab:res_japanese} and Table \ref{tab:res_aya23}, fine-tuning further on our IFT dataset significantly enhances the performance of these non-English-focused LLMs.

\section{Related Work}
\smallskip
\noindent\textbf{Multilingual LLMs.}
LLMs~\cite{brown2020language,chowdhery2023palm,touvron2023llama,gpt-4} have achieved remarkable results on various NLP tasks~\cite{hendrycks2020measuring,srivastava2022beyond}.
With over 7,000 languages spoken worldwide and approximately 2,500 classified as low-resource by ~\citet{joshi-etal-2020-state}, which are spoken by more than 1 billion people, there is a growing need to expand the language coverage of LLMs.
To develop LLMs with multilingual capabilities, one straightforward approach is to pretrain them on a diverse set of languages. For example, BLOOM~\cite{workshop2023bloom} is pretrained on 46 languages and 13 programming languages, while Llama-2~\cite{touvron2023llama} is pretrained primarily on English with additional data from 27 other languages.
Despite these efforts, the language coverage of these models remains limited and predominantly focused on English.
Another approach is to continually train LLMs with additional languages~\cite{cui2023efficient,basile2023llamantino,imanigooghari-etal-2023-glot500}.
In particular, Chinese-Llama~\cite{cui2023efficient} continually trains Llama on Chinese corpora and integrates additional Chinese tokens into the original vocabulary to further improve the Chinese ability.

\smallskip
\noindent\textbf{Instruction Tuning.}
Instruction tuning has been a key paradigm for LLMs to improve their general performance and ability to follow instructions~\cite{wei2021finetuned,wang-etal-2022-super,ding-etal-2023-enhancing}.
However, these models are predominantly tuned using English, resulting in significant discrepancies in performance across languages~\cite{huang-etal-2023-languages,etxaniz2023multilingual}.
Multilingual instruction tuning has effectively narrowed this performance gap~\cite{kew2023turning,chen-etal-2024-monolingual}.
Typically, the data for multilingual instruction tuning is derived through translation from English data~\cite{li2023bactrian,zhang2023bayling,2024aya}, but this approach often misses cultural nuances and can introduce unnatural responses.
Some efforts~\cite{2024aya} utilize templates to automatically create large amounts of multilingual data, but this method is constrained by limited diversity in the instructions.
We propose to generate instructions directly from original multilingual responses, which preserves the naturalness of responses and enhances the diversity of instructions.

\section{Conclusion}
In conclusion, our research addresses the notable disparity in Instruction Fine-Tuning (IFT) datasets, predominantly centered on English, by proposing a novel method for collecting multilingual IFT datasets. By leveraging English-focused LLMs and monolingual corpora, our approach maintains the naturalness of specific languages and ensures diversity in the datasets. The quality control through a scoring function further enhances the effectiveness of the generated datasets.

Our extensive experiments on generative tasks demonstrate that the models trained with our multilingual IFT datasets significantly outperform those trained on traditional translated and templated datasets. Moreover, our models show substantial improvements in discriminative tasks, indicating a better comprehension of language.

These results underscore the importance of diverse and high-quality multilingual datasets in enhancing the performance of large language models across various languages. Our method provides a viable solution to the challenges faced in creating effective multilingual IFT datasets, paving the way for more inclusive and capable language models. Future research can build upon this approach to further refine and expand the capabilities of LLMs in a broader range of linguistic contexts.

\section*{Limitations}
Since the instructions were generated by LLMs, there may be inherent biases originating from the underlying models used in this study. Nevertheless, the models used are open-source, extensively utilized by the community, and trained with the goals of reducing bias and enhancing safety and usefulness.

This study aims to systematically assess the effectiveness of generated instructions for given responses in various languages. Due to limitations in computing resources, we were unable to extend the proposed data creation framework beyond four languages. However, we endeavored to cover low, medium, and high-resource languages and evaluated our approach on several NLP tasks.

In our evaluation of LLMs using different IFT-style datasets, we selected two generative tasks and one discriminative task to demonstrate the impact of our dataset. The study was limited to three tasks due to computational and time constraints. However, these tasks are popular and widely used in evaluating multilingual LLMs.

In future work, we plan to extend our evaluation to LLMs optimized for additional languages and explore multiple benchmarks within each language to better understand the native aspects of LLM performance.

\bibliography{custom}

\appendix
\section{Prompt  details}

The task prompts are inspired by several NLP tasks. A few of the sample prompts are shown in Table \ref{tab:sample_task_prompts}. The scoring prompt used to evaluate our generated instruction and given response pair is given in Table \ref{tab:score_prompt}.
\begin{table}[htbp!]
    \centering
    \rowcolors{1}{gray!25}{white}
    \begin{tabularx}{0.5\textwidth}{|X|}
        \hline

        \rowcolor{gray!30}
        \texttt{
        \small
            \parbox{\dimexpr\linewidth-2\tabcolsep\relax}{%
                \vspace{5pt} 
                Response: \{\{response\}\} \\ \\ Given the above response, generate an appropriate instruction so that the given response can become an answer to the instruction. If required, include relevant context in the instruction. \\ \\ Instruction: \\
                \vspace{2pt} 
                }} \\
                
        \hline
        \rowcolor{gray!30}
        \texttt{
        \small
            \parbox{\dimexpr\linewidth-2\tabcolsep\relax}{%
                \vspace{5pt} 
                Response:\{\{response\}\}\\ \\ Given the above response, generate a question along with a related context so that by using these two the given response becomes a correct answer to the question.\\ \\ Question with context: \\
                \vspace{2pt} 
                }} \\
                
        \hline
        
        \rowcolor{gray!30}
        \texttt{
        \small
            \parbox{\dimexpr\linewidth-2\tabcolsep\relax}{%
                \vspace{5pt} 
                Response:\{\{response\}\} \\ \\ Given the above response, generate a longer text related to the response so that the given response is a summary of that longer text. \\ \\ Longer Text: \\
                \vspace{2pt} 
                }} \\
        \hline

        \rowcolor{gray!30}
        \texttt{
        \small
            \parbox{\dimexpr\linewidth-2\tabcolsep\relax}{%
                \vspace{5pt} 
                Response:\{\{response\}\} \\ \\ Given the above response, generate a question, context related to the response if required, four choices where one of the choices is the same as the given response and correct answer. Ensure that the given response is a correct answer to the question. Also, ensure that the choices are relevant to the question and are not too similar to each other. Please number the choices from A to D. Also output the correct choice at the end. \\ \\ Question: \\ \\
                A.  \\ \\
                B. \\ \\ 
                C. \\ \\ 
                D. \\ \\
                Answer: \\ \\
                \vspace{2pt} 
                }} \\
        \hline
        \rowcolor{gray!30}
        \texttt{
        \small
            \parbox{\dimexpr\linewidth-2\tabcolsep\relax}{%
                \vspace{5pt} 
                Response:\{\{response\}\} \\ \\ Given the above response, generate a math problem so that the given response is the correct answer to the math problem. \\ \\ Math Problem: \\ \\ 
                \vspace{2pt} 
                }} \\
        \hline
        
    \end{tabularx}
    \caption{Sample task prompts $\mathcal{P}_I$ used to generate instruction $\mathcal{I}_{en}$ in Equation \ref{eq:instruct_gen}.} 
\label{tab:sample_task_prompts}
\end{table}

\begin{table}[htbp!]
    \centering
    \rowcolors{1}{gray!25}{white}
    \begin{tabularx}{0.5\textwidth}{|X|}
        \hline

        \rowcolor{gray!30}
        \texttt{
        \small
            \parbox{\dimexpr\linewidth-2\tabcolsep\relax}{%
                \vspace{5pt} 
                Below is an instruction from a user and a candidate response. Evaluate whether or not the response is a good example of how an AI Assistant should respond to the user’s instruction. Assign a score using the following 5-point scale: \\
                1: The response is incomplete, vague, off-topic, controversial, or not exactly what the user asked for. It may miss content, start the numbered list incorrectly, or repeat the user's instruction. The response may come from another person's perspective, contain personal experiences, or include promotional or irrelevant text.\\
                2: The response addresses most of the user's requests but does not directly fulfill the instruction. It might provide a high-level methodology instead of an exact solution. \\
                3: The response is helpful, addressing all the basic asks from the user. It is complete and self-contained but written from another person's perspective rather than an AI assistant’s. It may include personal experiences, opinions, or references to comments sections and social media. \\
                4: The response is written from an AI assistant’s perspective, clearly focused on the instruction. It is complete, clear, comprehensive, well-organized, self-contained, and written in a helpful tone. Minor improvements could make it more concise and focused. \\
                5: The response is perfect, with a clear focus on being a helpful AI Assistant. It addresses the user's instruction without irrelevant sentences, providing high-quality content that demonstrates expert knowledge. It is very well written, logical, easy to follow, engaging, and insightful. \\
                Please provide a brief reasoning for your rating and then write "Score: <rating>" on the last line. \\
                Instruction: {{instruction}} \\
                Response: {{response}}
                \vspace{2pt} 
                }} \\
                
        \hline
        
    \end{tabularx}
    \caption{Scroing prompt $\mathcal{P}_s$ used in Equation \ref{eq:score} to evaluate the quality of a generated instruction and given response pair in the dataset curation phase. } 
\label{tab:score_prompt}
\end{table}

\section{Examples}

\begin{figure*}
    \centering
  \includegraphics[width=1.0\linewidth]{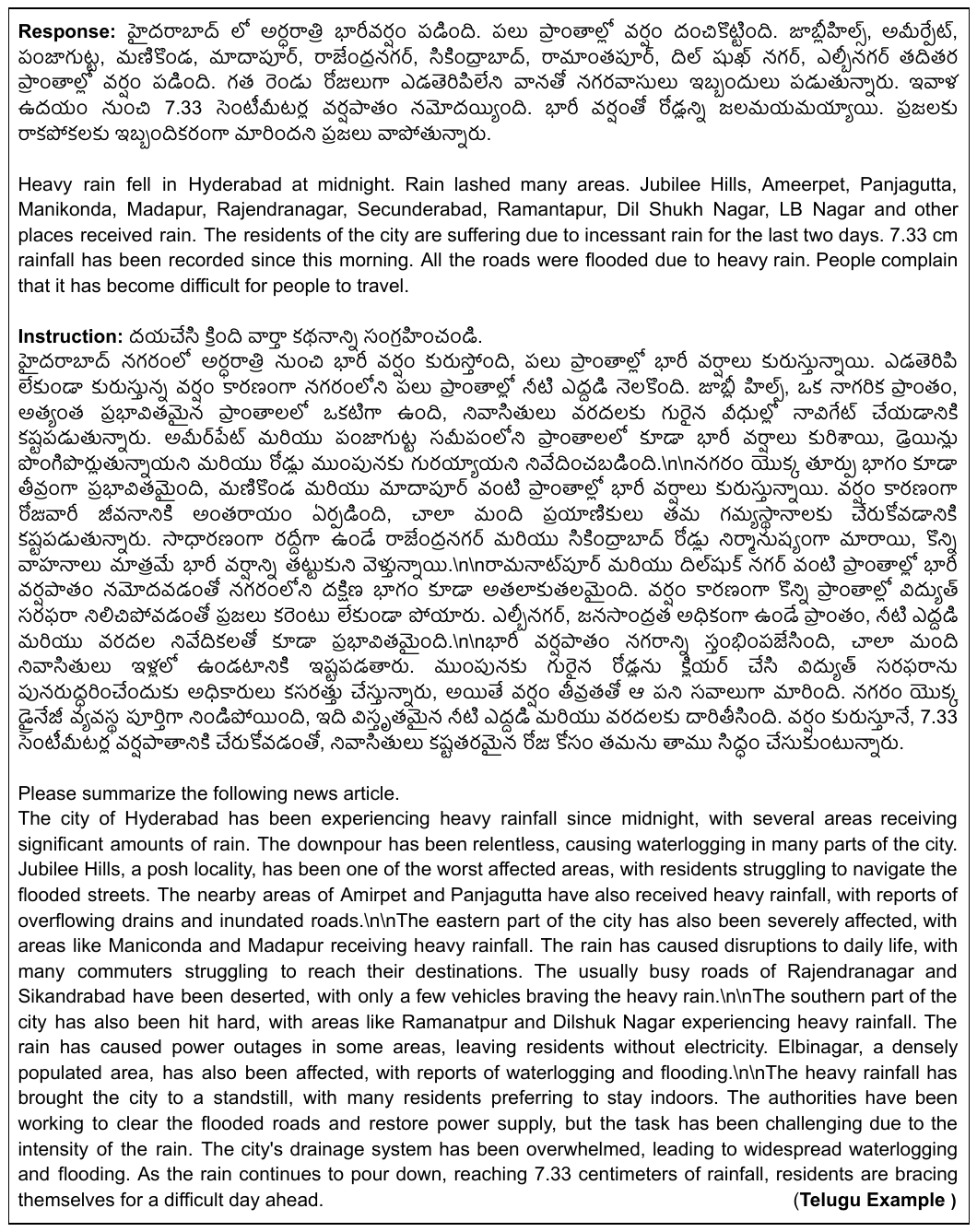}
  \vspace{-1cm}
    \caption{Telugu example based on summarization task}
    \label{fig:te_ex_1}
\end{figure*}



\begin{figure*}
    \centering
  \includegraphics[width=1.0\linewidth]{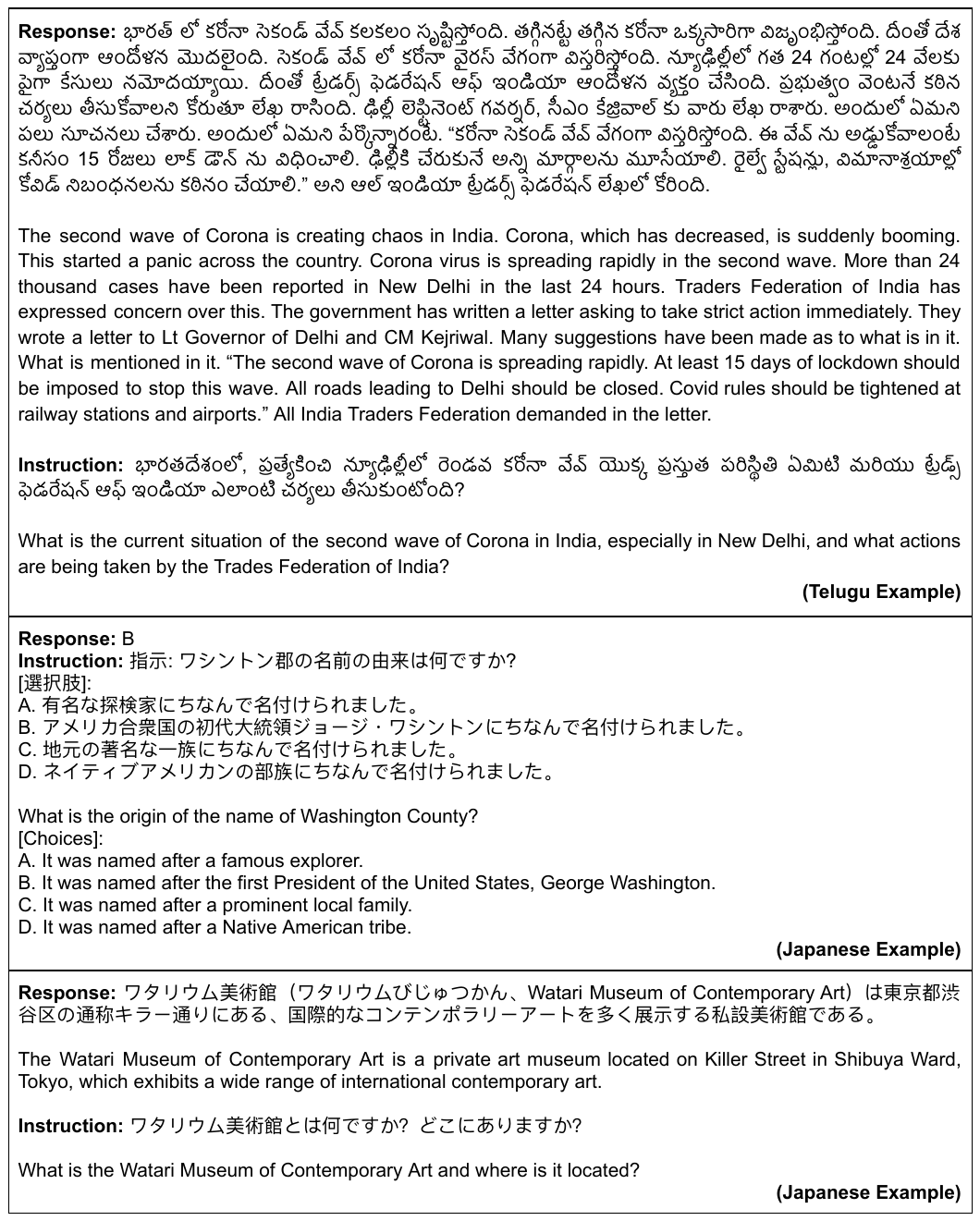}
   \vspace{-2cm}
    \caption{Japanese and Telugu examples based on question-answering tasks.}
    \label{fig:te_ex_2}
\end{figure*}

A few examples of multilingual IFT datasets created using our approach are shown in Figure \ref{fig:te_ex_1}, Figure \ref{fig:te_ex_2}. 

\section{Templated and Translated Datasets}
\begin{table*}[t]
    \centering
    \begin{tabular}{|c|c|c|c|} \hline
         \multirow{2}{*}{\textbf{Approach}} & \multicolumn{3}{c|}{Aya\_collection} \\
                                      & Human-annotation (\%)     &  Template datasets (\%) & Translation datasets (\%) \\ \hline
         Translation & 10 & 20 & 70 \\
         Template & 20 & 50 & 30 \\ \hline
         
    \end{tabular}
    \caption{Data sampling with different weighting schemes to create IFT datasets for translation-based and template-based approaches as described in \cite{2024aya}.} 
    \label{tab:aya_collection}
\end{table*}
The templated and translated datasets in aya are constructed using \textit{Aya\_dataset}, \textit{Aya\_collection} datasets by following the ratios described in Table \ref{tab:aya_collection}. The \textit{Aya\_dataset} is created by using native speakers from each language and contains approximately 6k examples per language. The \textit{Aya\_collection} is created by templating existing NLP datasets of each language as well as translating 19 datasets covering 93 languages. In total \textit{Aya\_collection} includes 513 million instances making it the largest open-source multilingual IFT dataset. For our experiments, we collected templated and translated datasets from \textit{Aya\_collection} for \textit{Telugu, Hindi, Chinese, and Spanish} languages.

\end{document}